\documentclass[a4paper,conference]{IEEEtran}

\ifCLASSINFOpdf
  \usepackage[pdftex]{graphicx}
\else
  \usepackage[dvips]{graphicx}
\fi

\usepackage{multirow}
\newcommand{\tabincell}[2]{\begin{tabular}{@{}#1@{}}#2\end{tabular}}
\usepackage{subfigure}
\usepackage{fmtcount}
\usepackage{color}

\begin{document}

\title{A Novel Integrated Framework for Learning both \\Text Detection and Recognition}

\author{

\IEEEauthorblockN{Wanchen Sui }
\IEEEauthorblockA{Alibaba\\
wanchen.swc@alibaba-inc.com}
\and
\IEEEauthorblockN{Qing Zhang }
\IEEEauthorblockA{Alibaba\\
sensi.zq@alibaba-inc.com}
\and
\IEEEauthorblockN{Jun Yang}
\IEEEauthorblockA{Alibaba\\
muzhuo.yj@alibaba-inc.com}
\and
\IEEEauthorblockN{Wei Chu }
\IEEEauthorblockA{Ant Financial, Alibaba\\
weichu.cw@alibaba-inc.com}

}

\maketitle

\begin{abstract}
In this paper, we propose a novel integrated framework for learning both text detection and recognition. For most of the existing methods, detection and recognition are treated as two isolated tasks and trained separately, since parameters of detection and recognition models are different and two models target to optimize their own loss functions during individual training processes. In contrast to those methods, by sharing model parameters, we merge the detection model and recognition model into a single end-to-end trainable model and train the joint model for two tasks simultaneously. The shared parameters not only help effectively reduce the computational load in inference process, but also improve the end-to-end text detection-recognition accuracy.
In addition, we design a simpler and faster sequence learning method for the recognition network based on a succession of stacked convolutional layers without any recurrent structure, this is proved feasible and dramatically improves inference speed.
Extensive experiments on different datasets demonstrate that the proposed method achieves very promising results.
\end{abstract}

\IEEEpeerreviewmaketitle

\section{Introduction}
Text in images provides rich and precise high-level semantic information, which is important for numerous potential applications such as scene understanding, image and video retrieval, and content-based recommendation systems.
Therefore, considerable efforts 
have been made towards the automatic detection and recognition of text in images, which is still an important challenge for visual understanding \cite{jaderberg2016reading}.

As stated in \cite{yao2014unified}, detection and recognition components in conventional scene text recognition systems are sequential and isolated in the whole system pipeline, this makes it difficult to exploit interactive information between the two components. 
Several works proposed uniformed frameworks for text detection and recognition.
Li \emph{et al.} \cite{iccv2017} presented a unified network that simultaneously localizes and recognizes text with a single forward pass, based on convolutional recurrent neural networks.
Bartz \emph{et al.} \cite{stn-ocr} designed a single deep neural network to learn to detect and recognize text from natural images in a semi-supervised way.
Bu{\v{s}}ta \emph{et al.} \cite{text-spotter} proposed an end-to-end trainable scene text localization and recognition framework called Deep TextSpotter.
Different from these methods, our system can perform text detection and recognition via a single framework based entirely on Convolution Neural Network (CNN), and is proved to perform well in cursive English handwriting recognition and Chinese text recognition.

In this work, we design an end-to-end trainable neural network to address this problem, condensing the detection network and recognition network into a single one, which can be trained in an end-to-end manner. By sharing the convolutional layers, we can compute the shared feature maps from entire input image only once, as a result, the text recognition system can be effectively accelerated.
In order to make a further speedup for inference, we apply a succession of convolutional layers instead of recurrent layers for sequence learning task. Since our proposed recognition network is based entirely on convolutional neural networks, computations over all elements can be fully parallelized for inference optimization. Meanwhile, it is proved to perform on par with recurrent sequence learning method with respect to recognition accuracy.

In our work, we choose two scenarios to verify our proposed methods for end-to-end text recognition task. We first make a large collection of business cards and get manual labelling to develop a business card recognition system. In addition, we conduct extensive experiments on end-to-end handwritten document recognition using public IAM dataset \cite{iam} as benchmark.

In summary, the contributions of the paper are as follows: 

(1) A novel integrated framework is proposed for text detection and recognition, which can accomplish these two tasks in an end-to-end trainable neural network. 

(2) Given the  detection and recognition networks, sharing their common convolutional features can get speedup in recognition computation.

(3) The shared convolutional feature maps, which exploit interaction between two different tasks, can be effective in text detection and recognition and improve performance of end-to-end model.

(4) We propose a simpler and faster recognition network based on a succession of convolutional layers without any recurrent layer which is often used for sequence learning. In this way, by combining with Region-based Fully Convolutional Networks (R-FCN) \cite{rfcn}, which is a region-based object detection method, our system can perform text detection and recognition via a fully convolutional network.



\section{The Proposed Network Architecture} \label{section:method}
In this section, we present our main ideas and details of the proposed algorithm. Specifically, we give an overview of the proposed framework in Section \ref{section:overview}, explain the text detection network and text recognition network in Section \ref{section:detection} and Section \ref{section:recognition}, illustrate the proposed text pooling layer in Section \ref{section:text-pool}, and describe the loss function and implementation details in Section \ref{section:loss}.  

\subsection{Overview} \label{section:overview}
\begin{figure}
\centerline{\includegraphics[height=2.0in,width=3.5in]{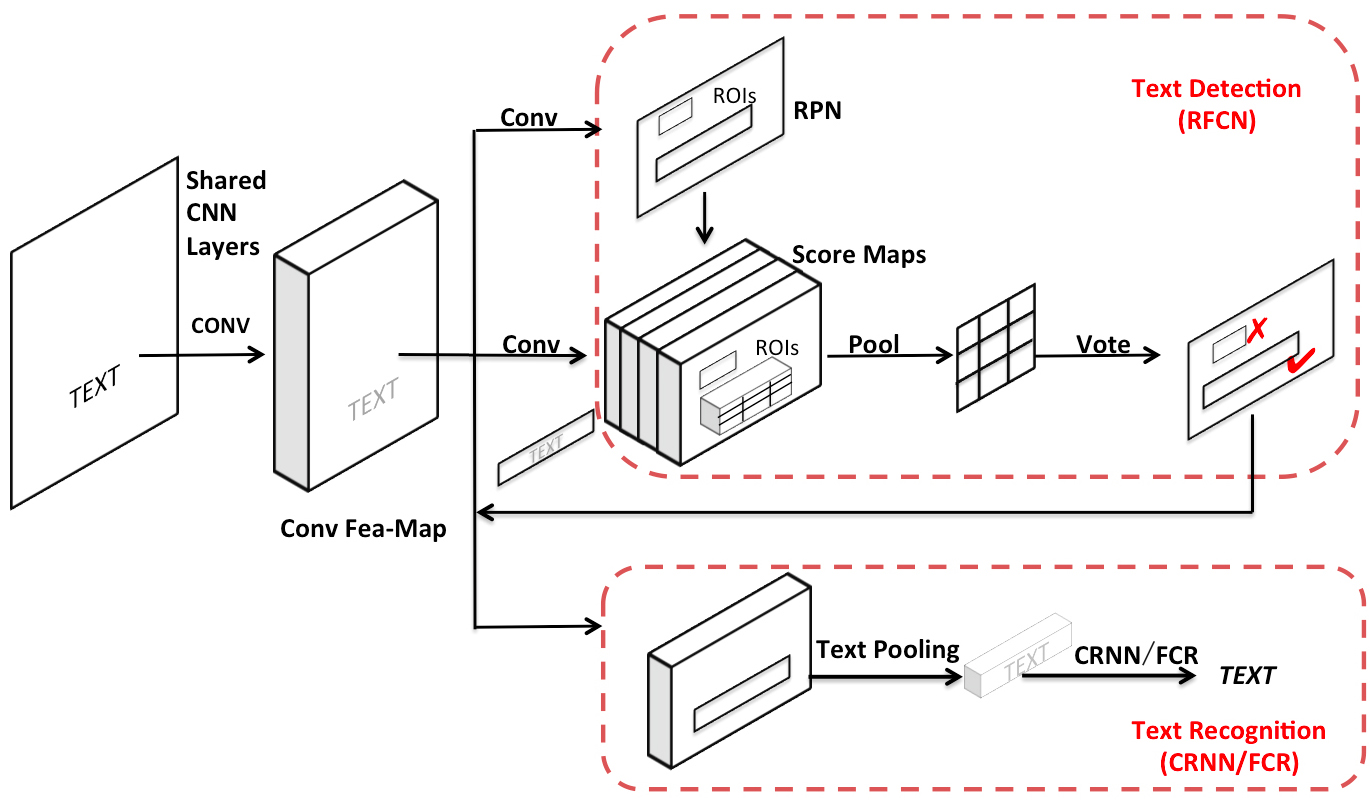}}
\caption{Schematic overview of the proposed framework}
\label{fig:overview}
\end{figure}
A schematic overview of the proposed framework is illustrated in Figure \ref{fig:overview}.
Firstly, the whole image is processed with several convolutional layers and max pooling layers to generate convolutional feature maps, which are shared by text detector and recognizer. The network uses the very deep VGG-16 network \cite{vgg16} as its backbone (conv1$\_$1 through conv4$\_$3), while other networks such as Googlenet \cite{googlenet}, Resnet \cite{resnet} and so on 
are also applicable.

Then we perform text detection using a CNN based detection method. In this paper,  we take R-FCN \cite{rfcn} as an example to describe our approach, which is readily applicable to other CNN based detection method, such as Faster Region-based CNN (R-CNN) \cite{faster-rcnn}, You Only Look Once (YOLO) \cite{yolo}, Single Shot MultiBox Detector (SSD) \cite{ssd}. The detector returns the regions containing individual lines of text.
In contrast to most existing algorithms that directly perform text recognition on each cropped text line of the input image, we design a text line pooling layer similar to Region of Interest (RoI) pooling layer \cite{fast-rcnn}, to extract feature maps of a fixed height from the aforementioned shared convolutional feature maps for each detected text line. Then, these small text-line feature maps are sent to a text recognition network.

For the text recognizer,  we propose two different schemes to perform text recognition based on feature maps of each extracted text-line. One is a sequence recognition method
Convolutional Recurrent Neural Network (CRNN) \cite{crnn},
a combination of Deep Convolutional Neural Networks (DCNN), Recurrent Neural Networks (RNN) layer and Connectionist Temporal Classification (CTC). In the other scheme, we design a simpler and faster recognition network, applying several successive convolutional layers instead of the recurrent layers in CRNN. We denote the newly designed network as Fully Convolutional Recognizer (FCR).

To summarize, our network takes an entire image as input, extracts convolutional features from input only once, makes feature sharing for both detection and recognition, and finally outputs the detected text regions and the corresponding text contents. 

\subsection{Text Detection Network} \label{section:detection}
In this paper, we deploy a region proposal based object detection method named R-FCN to perform text detection. 
R-FCN adopts a popular two-stage object detection strategy: region proposal and proposal classification.
Correspondingly, in this detection network, the aforementioned shared convolutional feature maps finally branch into two sibling sub-networks: the R-FCN network uses a Region Proposal Network (RPN) for extracting candidate regions, and ends with a position-sensitive RoI pooling layer for generating scores and refining coordinates for each proposal.

R-FCN is designed to preform accurate and efficient object detections. Nevertheless, text regions are usually of their specific shapes which are different from that of general objects. In our framework, we modify the specific settings in the R-FCN architecture to cover a wide range of text lines: the scales of anchors are set to 4, 8 and 16, and the aspect ratios are changed to 1:2, 1:5 and 1:10.
\subsection{Text Pooling Layer} \label{section:text-pool}
To handle the detected text regions, we append another sub-network over the convolutional feature map outputs of the last shared convolutional layer. At the bottom of the sub-network, a text pooling layer is designed to extract a fixed-height feature maps for individual text region. 

Specially, it uses max pooling to reshape the shared feature maps inside a valid text region to fixed height $H$, keeping the original height/width ratios. Here, $H$ is the layer hyper-parameter, which is constant for any particular regions. In this work, the detected text regions are horizontal rectangular bounding boxes. The text pooling layer divides an $h \times w$ text region into $H \times \frac{w \times H}{h}$ sub-regions, the size of each grid is approximate $\frac{h}{H} \times \frac{h}{H}$, and the values in each sub-region are downsampled to the corresponding output grid cell by max-pooling. Each feature map channel processes pooling independently, which is the same as the standard max pooling.

In virtue of the text pooling layer, the network is able to jointly optimize the detector and recognizer with an end-to-end training strategy. Additionally, the end-to-end system computes the shared feature maps from the entire image only once, this can effectively reduce computational load during inference process.

\subsection{Text Recognition Network} \label{section:recognition}
As mentioned above, the text regions are converted into fixed-height feature maps via a text pooling layer, and finally sent to a sequence recognizer to predict the corresponding text contents.
As for the recognizer, we adopt two different schemes to perform text recognition based on feature maps of each extracted text-line.

The first one is a sequence recognition method named CRNN \cite{crnn}, which integrates feature extraction, sequence modeling and transcription into an end-to-end trainable network.
In our framework, we take advantage of the shared convolutional layers to process feature extraction. Thus, these small text-line feature maps can be directly fed into sequence modeling and transcription layers.
To be specific, we deploy a Bidirectional Recurrent Neural Network (BiLSTM) to predict a label distribution for each frame, and then translate the per-frame predictions into the final text sequence.

Recently, Gehring \emph{et al}. \cite{facebook} proposed a fully convolutional model for sequence to sequence learning, which outperforms recurrent models on very large benchmark datasets at an order of magnitude faster speed.
Inspired by this work, we design a simpler and faster recognition network named FCR, using a succession of convolutional layers in place of the BiLSTM layers in CRNN. Through combining R-FCN and FCR, our system can perform text detection and recognition via a single network that entirely consists of convolutional layers.

\subsection{Loss Function} \label{section:loss}
In this work, we employ multi-task learning to jointly optimize model parameters. According to the multi-task loss applied in Faster RCNN, we define the overall loss function for each image as:
\begin{equation}\label{eq:loss}
    L=\sum_iL_i^{det}+\lambda \sum_jL_j^{ctc},
\end{equation}
where $i$ is the index of a proposal in a mini-batch for detection, $j$ is the index of a text line in a mini-batch for recognition, $L_i^{det}$ is the detection loss on the $i$-th proposal, $L_j^{ctc}$ is the CTC \cite{ctc} loss on the $j$-th labeled text line, and $\lambda>0$ is the trade-off parameter.

Following R-FCN, the first loss $L^{det}$ on each proposal is defined as:
\begin{equation}\label{eq:loss_det}
    L^{det}=L^{cls}+\gamma c^* L^{reg},
\end{equation}
where $L^{cls}$ is the cross-entropy loss for classification, $L^{reg}$ is the bounding box regression loss, $c^*$ is the ground-truth label of the proposal ($c^*=1$ means positive proposal, $c^*=0$ means negative one), $\gamma$ is the balance weight.


\section{Experiment} \label{section:experiment}
\subsection{Datasets} \label{section:datasets}
\textbf{Chinese Business Card Database} consists of 20 thousands Chinese business card photographs taken from different viewpoints under varying illumination conditions. It contains about 200 thousands text lines with a dictionary of 3800 characters. The business cards with different size, color and font are collected and labeled manually, we uniformly sample 16 thousands text lines as training data, the rest as testing data.

\textbf{IAM Handwriting Database} \cite{iam} is an English sentence database for offline handwriting recognition. It contains forms of unconstrained handwritten text, which are scanned at a resolution of 300dpi and saved as PNG images with 256 gray levels. 
All texts in the database are built using sentences provided by the Lancaster-Oslo/Bergen(LOB) Corpus. 657 writers produced between 1 and 59 handwritten documents in their own handwriting. 
There are 747 documents (6,161 lines) in the training set, 105 documents (900 lines) in the validation1 set, 115 documents (940 lines) in the validation2 set and 232 documents (1,861 lines) in the test set.
The texts in this database typically contain 450 characters in about nine lines.

\subsection{Setup} \label{section:setup}
In the experiments,  we adopt the VGG-16 network as backbone of the network, conv1$\_$1, conv1$\_$2, conv2$\_$1, conv2$\_$2, conv3$\_$1, conv3$\_$2, conv3$\_$3, conv4$\_$1, conv4$\_$2, conv4$\_$3 are selected to be shared between detection and recognition tasks. The network can be trained end-to-end using the standard error back-propagation and ADAM optimizer \cite{adam}. In the training process, we set the balance weight $\lambda=1$ in formula (\ref{eq:loss}) and $\gamma=1$ in formula (\ref{eq:loss_det}). We perform two different training strategies:  separate training and joint training. For separate training, the shared layers are initialized and fixed by pre-training a model for ImageNet classification, the rest layers (include the non-shared convolutional layers and the typical layers for detector and recognizer) are separately trained by the detector and recognizer. For joint training, we employ multi-task learning to jointly optimize model parameters. The shared layers are jointly trained by the detector and recognizer, and the whole network is trained in an end-to-end strategy using the standard error back-propagation.

We use these two training strategies to share different number of convolutional layers based on the Chinese Business Card database. For each strategy, a baseline method (share no layer) is compared with the proposed models. For Business Card database, a bounding box is considered as correct only if its Intersection over Union (IoU) ratio with any ground-truth is greater than 0.5 and the recognized character sequence also matches.
In order to validate the performance further, we carry out experiments on a public database, IAM database, and compare our best model with the existing systems. For IAM database, the measurement of performance is the Word Error Rate (WER\%), corresponding to the edit distance between the recognition result and ground-truth, normalized by the number of answers in ground-truth.

\subsection{Evaluation Under Different Settings}
Table \ref{tab:rfcn_lstm_fix} illustrates the results of our method using the combination of R-FCN and CRNN on Chinese Business Card database by separately training. 
The first column represents the shared layers, ``None'' indicating no layer is shared. The second column ``Detection" gives text detection results, including Recall, Precision and Average Precision (AP). The column ``Recognition" lists sequence and character accuracies given the ground-truth text regions. The last column shows the end-to-end recognition results, including Recall, Accuracy and F-measure at line level.
It can be seen that if we make convolutional layers shared until conv2$\_$2, the end-to-end system can get improvement for both precision and computational efficiency.
We also report the results of our method using R-FCN and FCR on Chinese Business Card database in Table \ref{tab:rfcn_fcn_fix} , which obtain the similar trend.


\begin{table*}[ht!]
\centering
\caption{Results of our method using the combination of R-FCN and CRNN on Chinese Business Card database by separately training.}
\label{tab:rfcn_lstm_fix}
\begin{tabular}{|c|c|c|c|c|c|c|c|c|}
\hline
    \multirow{2}{*}{Shared Layers}
    & \multicolumn{3}{|c|}{Detection (R-FCN)}
    & \multicolumn{2}{|c|}{Recognition (CRNN)}
    & \multicolumn{3}{|c|}{End-to-End} \\
    \cline{2-9}
    & Recall & Precision & AP & Seq Acc & Char Acc & Recall & Accuracy & F-measure\\
\hline
    None                       & 0.9723 & 0.8412 & 0.9485 & 0.8442 & 0.9799 & 0.7193 & 0.6755  & 0.6967 \\
\hline
    conv1\_1 -- conv1\_2 & 0.9629 & 0.8577 & 0.9414 & 0.8397 & 0.9792 & 0.7233 & 0.6993 & 0.7110 \\
\hline
    conv1\_1 -- conv2\_2 & 0.9660 & 0.8480 & 0.9423 & 0.8310 & 0.9777 & \color{blue} \textbf{0.7223} & \color{blue} \textbf{0.6883} & \color{blue} \textbf{0.7048} \\
\hline
    conv1\_1 -- conv3\_3 & 0.9618 & 0.8552 & 0.9388 & 0.7911 & 0.9703 & 0.6802 & 0.6565 & 0.6881 \\
\hline
    conv1\_1 -- conv4\_3 & 0.9563 & 0.8544 & 0.9313 & 0.6470 & 0.9387 & 0.5557 & 0.5389 & 0.5471 \\
\hline
\end{tabular}
\end{table*}

\begin{table*}[ht!]
\centering
\caption{Results of our method using the combination of R-FCN and FCR on Chinese Business Card database by separately training.}
\label{tab:rfcn_fcn_fix}
\begin{tabular}{|c|c|c|c|c|c|c|c|c|}
\hline
    \multirow{2}{*}{Shared Layers}
    & \multicolumn{3}{|c|}{Detection (R-FCN)}
    & \multicolumn{2}{|c|}{Recognition (FCR)}
    & \multicolumn{3}{|c|}{End-to-End} \\
    \cline{2-9}
    & Recall & Precision & AP & Seq Acc & Char Acc & Recall & Accuracy & F-measure\\
\hline
    None                       & 0.9723 & 0.8412 & 0.9485 & 0.8376 & 0.9780 & 0.7129 & 0.6695  & 0.6905 \\
\hline
    conv1\_1 -- conv1\_2 & 0.9703 & 0.8449 & 0.9486 & 0.8407 & 0.9782 & 0.7302 & 0.6902 & 0.7096 \\
\hline
    conv1\_1 -- conv2\_2 & 0.9708 & 0.8427 & 0.9491 & 0.8273 & 0.9759 & \color{blue} \textbf{0.7213} & \color{blue} \textbf{0.6796} & \color{blue} \textbf{0.6998} \\
\hline
    conv1\_1 -- conv3\_3 & 0.9681 & 0.8492 & 0.9466 & 0.7945 & 0.9693 & 0.7008 & 0.6673 & 0.6837 \\
\hline
    conv1\_1 -- conv4\_3 & 0.9536 & 0.8477 & 0.9288 & 0.5574 & 0.9121 & 0.4867 & 0.4696 & 0.4780 \\
\hline
\end{tabular}
\end{table*}

Table \ref{tab:rfcn_lstm} and Table \ref{tab:rfcn_fcn} show the joint training results, from which we can observe that: 

(1) The models with a certain number of shared layers can achieve better performance than the baseline (``None", non-shared model). For the combination of R-FCN and CRNN, sharing convolutional layers from ``conv1$\_$1'' to ``conv2$\_$2" performs best. Employing the FCR, sharing convolutional layers from ``conv1$\_$1" to ``conv3$\_$3" yields the highest F-measure.

(2) Compared with the separate training results, the shared layers contribute to the end-to-end recognition improvement by joint training. The knowledge learned from the correlated detection and recognition tasks may be effectively transferred between each other.

(3) The models with different numbers of shared layers have different performance. If sharing too many convolutional layers, the recognition accuracy may decline to some extent, such as sharing layers from ``conv1$\_$1" to ``conv4$\_$3" using R-FCN and CRNN. 

\begin{figure}
\centerline{\includegraphics[height=1.8in]{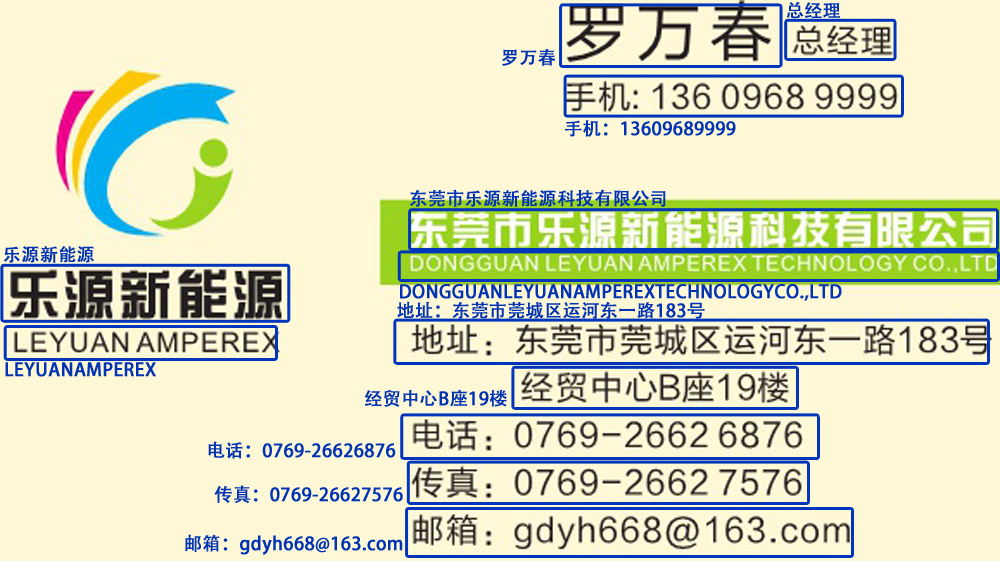}}
\caption{Examples of end-to-end text recognition results.}
\label{fig:results}
\end{figure}

Several end-to-end recognition examples are illustrated in Figure \ref{fig:results}.

\begin{table*}[ht!]
\centering
\caption{Results of our method using the combination of R-FCN and CRNN on Chinese Business Card database by jointly training.}
\label{tab:rfcn_lstm}
\begin{tabular}{|c|c|c|c|c|c|c|c|c|}
\hline
    \multirow{2}{*}{Shared Layers}
    & \multicolumn{3}{|c|}{Detection (R-FCN)}
    & \multicolumn{2}{|c|}{Recognition (CRNN)}
    & \multicolumn{3}{|c|}{End-to-End} \\
    \cline{2-9}
    & Recall & Precision & AP & Seq Acc & Char Acc & Recall & Accuracy & F-measure\\
\hline
    None                       & 0.9723 & 0.8412 & 0.9485 & 0.8442 & 0.9799 & 0.7193 & 0.6755  & 0.6967 \\
\hline
    conv1\_1 -- conv1\_2 & 0.9668 & 0.8521 & 0.9449 & 0.8447 & 0.9800 & 0.7252 & 0.6938 & 0.7092 \\
\hline
    conv1\_1 -- conv2\_2 & 0.9643 & 0.8600 & 0.9425 & 0.8388 & 0.9791 & \color{blue} \textbf{0.7251} & \color{blue} \textbf{0.7020} & \color{blue} \textbf{0.7134} \\
\hline
    conv1\_1 -- conv3\_3 & 0.9638 & 0.8674 & 0.9442 & 0.8325 & 0.9780 & \color{blue} \textbf{0.7212} & \color{blue} \textbf{0.7045} & \color{blue} \textbf{0.7128} \\
\hline
    conv1\_1 -- conv4\_3 & 0.9605 & 0.8623 & 0.9424 & 0.7976 & 0.9735 & 0.6734 & 0.6562 & 0.6647 \\
\hline
\end{tabular}
\end{table*}

\begin{table*}[ht!]
\centering
\caption{Results of our method using the combination of R-FCN and FCR on Chinese Business Card database by jointly training.}
\label{tab:rfcn_fcn}
\begin{tabular}{|c|c|c|c|c|c|c|c|c|}
\hline
    \multirow{2}{*}{Shared Layers}
    & \multicolumn{3}{|c|}{Detection (R-FCN)}
    & \multicolumn{2}{|c|}{Recognition (FCR)}
    & \multicolumn{3}{|c|}{End-to-End} \\
    \cline{2-9}
    & Recall & Precision & AP & Seq Acc & Char Acc & Recall & Accuracy & F-measure\\
\hline
    None                       & 0.9723 & 0.8412 & 0.9485 & 0.8376 & 0.9780 & 0.7129 & 0.6695  & 0.6905 \\
\hline
    conv1\_1 -- conv1\_2 & 0.9710 & 0.8433 & 0.9492 & 0.8455 & 0.9793 & 0.7322 & 0.6903 & 0.7106 \\
\hline
    conv1\_1 -- conv2\_2 & 0.9711 & 0.8466 & 0.9503 & 0.8393 & 0.9779 & \color{blue} \textbf{0.7327} & \color{blue} \textbf{0.6934} & \color{blue} \textbf{0.7125} \\
\hline
    conv1\_1 -- conv3\_3 & 0.9712 & 0.8525 & 0.9508 & 0.8378 & 0.9779 & \color{blue} \textbf{0.7402} & \color{blue} \textbf{0.7052} & \color{blue} \textbf{0.7223} \\
\hline
    conv1\_1 -- conv4\_3 & 0.9665 & 0.8459 & 0.9448 & 0.8294 & 0.9768 & \color{blue} \textbf{0.7245} & \color{blue} \textbf{0.6883} & \color{blue} \textbf{0.7059} \\
\hline
\end{tabular}
\end{table*}

\subsection{Comparison to Published Results}
In this section, we perform experiments on IAM database and compare our models with the existing methods. Using the joint training strategy, we conduct experiments on combinations of R-FCN + CRNN and R-FCN + FCR, while the R-FCN can use VGG-16 or Resnet-50 \cite{resnet} as its backbone. Feature Pyramid Network (FPN) \cite{fpn} is applied to our recognizer as a generic feature extractor to improve the performance. The comparison of our experiments on IAM are shown in Table \ref{tab:iam_our}, we can see the Resnet-50 based R-FCN + CRNN/FCR framework achieves better recognition results.
Our best result is compared with reported results of existing methods in Table \ref{tab:iam_compare}. When comparing the error rates, it is important to note that methods \cite{bluche2015deep,doetsch2014fast,kozielski2013improvements,pham2014dropout} all used an explicit (ground-truth) line segmentation and a language model. Method \cite{nips2016} is an end-to-end handwritten paragraph recognition method, comparing to it, our system (using R-FCN and CRNN based on Resnet-50)  can achieve competitive result, which is slightly better in WER\%.

\begin{table}[ht!]
\centering
\caption{Results of our method on IAM database.}
\label{tab:iam_our}
\begin{tabular}{|c|c|c|c|}
\hline
    Method & Network & Shared Layers & WER\% \\
\hline
    \multirow{2}{*}{R-FCN+CRNN} & Resnet-50 & conv1 -- res2c & 24.4 \\
\cline{2-4}
    & VGG-16 & conv1\_1 -- conv2\_2 & 30.9 \\
\hline
    \multirow{2}{*}{R-FCN+FCR} & Resnet-50 & conv1 -- res3d & 28.8 \\
\cline{2-4}
    & VGG-16 & conv1\_1 -- conv3\_3 & 31.2 \\
\hline
\end{tabular}
\end{table}

\begin{table}[ht!]
\centering
\caption{Results of different approaches on IAM database.}
\label{tab:iam_compare}
\begin{tabular}{|c|c|c|c|}
\hline
    Mehods & Language Model & Line Segmentation & WER\% \\
\hline
    Pham \emph{et al}. \cite{pham2014dropout} & Yes & No & 13.6 \\
\hline
    Kozielski \emph{et al}. \cite{kozielski2013improvements} & Yes & No & 13.3 \\
\hline
    Doetsch \emph{et al}. \cite{doetsch2014fast} & Yes & No & 12.2 \\
\hline
    Bluche \cite{bluche2015deep} & Yes & No & 10.9 \\
\hline
    Bluche \cite{nips2016} & Yes & Yes & 16.4 \\
\hline
    Bluche \cite{nips2016} & No & Yes & 24.6 \\
\hline
    Our best result* & No & Yes & 24.4 \\
\hline
\end{tabular}
\end{table}

\subsection{Impact of Sharing Layers}

In deep convolution neural networks, with the increasing convolutional layers, each point in the corresponding feature map has larger receptive field on the input image. Taking VGG-16 as an example, the size of receptive fields for ``conv2$\_$2'' and ``conv3$\_$3'' feature maps are 14$\times$14 and 40$\times$40, respectively.

\begin{figure}
\centerline{\includegraphics[height=1.8in]{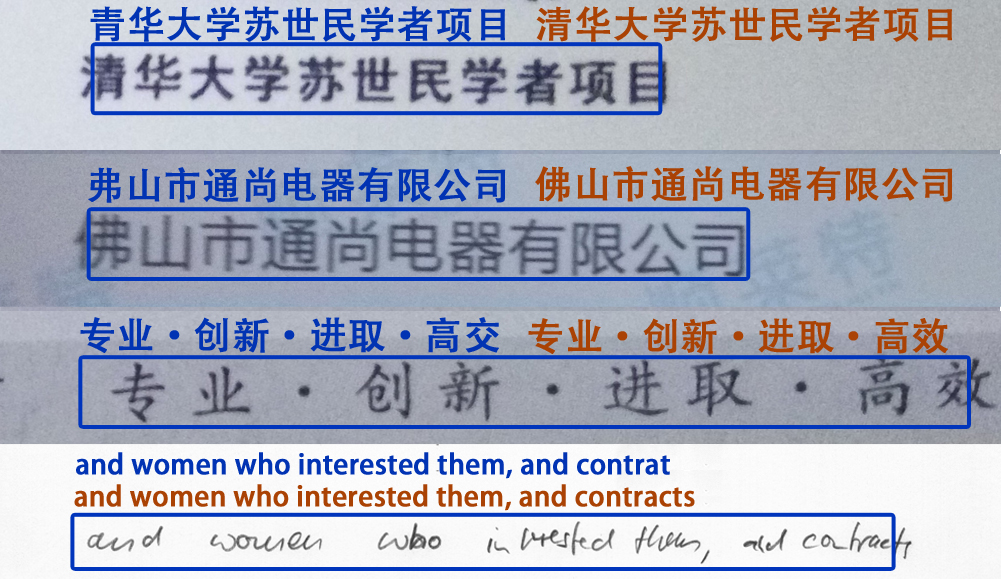}}
\caption{The comparison between the models with shared layers and without shared layers, in case that the bounding boxes do not cover exactly. The orange character sequences are the recognition result of models with shared layers, the blue results from models without shared layers.}
\label{fig:reception_result}
\end{figure}

In our framework, we use a text pooling layer to extract features from the feature map outputs of the last shared convolutional layer according to each detected region. And each point in the feature map has considerable receptive field on the input image. Thus, comparing to performing text recognition on each cropped text line of the input image, our method can obtain more effective information. It is important to note that if a detected box does not precisely cover the whole character area, directly cropping the detected region from input image for recognition will cause irreparable loss of information, especially the leftmost and rightmost boundary. In contrast, extracting from the feature map potentially acquires the information beyond the given bounding box and can contribute to improve the recognition performance, as shown in Figure \ref{fig:reception_result}.

\subsection{Inference Efficiency}
In this part, we evaluate inference efficiency of our architecture. 
All experiments are implemented on an NVIDIA Tesla M40 GPU with 12GB memory.
As illustrated in Table \ref{tab:rfcn_lstm_time} and Table \ref{tab:rfcn_fcn_time}, for the combination of R-FCN and CRNN, we can share convolutional layers from ``conv1$\_$1" to ``conv3$\_$3" without sacrificing the accuracy, this can make 53\% computational saving and achieve 1.56$\times$ speedup for text recognition. As for the combination of R-FCN and FCR, sharing layers from ``conv1$\_$1" to ``conv4$\_$3" acquires 75\% computational saving and  2.41$\times$ speedup in inference, while keeping the end-to-end result higher than baseline.
In addition, comparing the two tables, applying convolutional layers instead of BiLSTM in recognizer gains considerable speed improvements both theoretically and practically.
Table \ref{tab:rfcn_lstm_time_iam} shows the inference speed of our architecture on IAM database.

\begin{table*}[ht!]
\centering
\caption{The inference speed of our architecture using the combination of R-FCN and CRNN on Chinese Business Card database.}
\label{tab:rfcn_lstm_time}
\begin{tabular}{|c|c|c|c|c|}
\hline
    Shared Layers & R-FCN & CRNN & \tabincell{c}{Runtime speedup \\ for CRNN} & \tabincell{c}{Computational saving \\ for CRNN}\\
\hline
    None                       & \multirow{4}{*}{0.0888$\pm$0.0081} & 0.1494$\pm$0.0008 & - & - \\
\cline{1-1}\cline{3-5}
    conv1\_1 -- conv2\_2 & & 0.1166$\pm$0.0005 & 1.28$\times$ & 26.78\% \\
\cline{1-1}\cline{3-5}
    conv1\_1 -- conv3\_3 & & 0.0955$\pm$0.0003 & 1.56$\times$ & 53.41\% \\
\cline{1-1}\cline{3-5}
    conv1\_1 -- conv4\_3 & & 0.0804$\pm$0.0002 & 1.86$\times$ & 80.04\% \\
\hline
\end{tabular}
\end{table*}

\begin{table*}[ht!]
\centering
\caption{The inference speed of our architecture using the combination of R-FCN and FCR on Chinese Business Card database.}
\label{tab:rfcn_fcn_time}
\begin{tabular}{|c|c|c|c|c|}
\hline
    Shared Layers & R-FCN & FCR & \tabincell{c}{Runtime speedup \\ for FCR} & \tabincell{c}{Computational saving \\ for FCR}\\
\hline
    None                       &\multirow{4}{*}{0.0888$\pm$0.0081} & 0.0583$\pm$0.0005 & - & - \\
\cline{1-1}\cline{3-5}
    conv1\_1 -- conv2\_2 & & 0.0461$\pm$0.0004 & 1.26$\times$ & 25.15\% \\
\cline{1-1}\cline{3-5}
    conv1\_1 -- conv3\_3 & & 0.0368$\pm$0.0003 & 1.58$\times$ & 50.16\% \\
\cline{1-1}\cline{3-5}
    conv1\_1 -- conv4\_3 & & 0.0242$\pm$0.0002 & 2.41$\times$ & 75.17\% \\
\hline
\end{tabular}
\end{table*}

\begin{table*}[ht!]
\centering
\caption{The inference speed of our architecture on IAM database.}
\label{tab:rfcn_lstm_time_iam}
\begin{tabular}{|c|c|c|c|c|}
\hline
    Method & Shared Layers & R-FCN Detection & Recognition & \tabincell{c}{Runtime speedup \\ for Recognition} \\
\hline
    \multirow{2}{*}{\tabincell{c}{R-FCN+CRNN \\ (Resnet-50)}} & None & \multirow{2}{*}{0.2790$\pm$0.0023} & 0.3455$\pm$0.0036 & - \\
\cline{2-2}\cline{4-5}
    & conv1 -- res2c & & 0.2847$\pm$0.0057 & 1.21$\times$ \\
\hline
     \multirow{2}{*}{\tabincell{c}{R-FCN+FCR \\ (Resnet-50)}} & None & \multirow{2}{*}{0.2790$\pm$0.0023} & 0.2863$\pm$0.0019 & - \\
\cline{2-2}\cline{4-5}
    & conv1 -- res3d & & 0.1012$\pm$0.0003 & 2.83$\times$ \\
\hline
\end{tabular}
\end{table*}


\section{Conclusion} \label{section:conclusion}
We have presented an end-to-end trainable neural network, which solves text detection and recognition tasks in a novel integrated framework. By sharing the convolutional feature maps, we can simultaneously train these two models in a single unified pipeline. 
In addition, we designed a fully convolutional recognizer, which has been proved effective and efficient.
In this way, given the computation of detection network, recognition network can achieve about 50\% computation saving. 
Experiment results suggest that the proposed method is not only a cost-efficient solution for practical usage, but also an effective way of improving text detection and recognition accuracy.
In this framework, we mainly focus on horizontal or near-horizontal texts, as for future work, we will pay more attention to handling texts of multiple orientations.


\bibliographystyle{IEEEtran}
\bibliography{z_bibfile}

\end{document}